\documentclass[11pt,a4paper]{article}
\usepackage[hyperref]{conll2018}
\usepackage{times}
\usepackage{latexsym}

\usepackage{url}
\usepackage{xspace}
\usepackage{amsmath}
\usepackage{amsfonts}
\usepackage{booktabs}
\usepackage{algorithm}
\usepackage[noend]{algpseudocode}
\usepackage{amssymb}
\usepackage{array,multirow,graphicx}

\let\emptyset\varnothing

\usepackage{tikz}
\usetikzlibrary{bayesnet}

\aclfinalcopy 

\usepackage{cleveref}
\crefname{section}{\S}{\S\S}
\Crefname{section}{\S}{\S\S}
\crefname{table}{Tab.}{}
\crefname{figure}{Fig.}{}
\crefname{algorithm}{Alg.}{}
\crefname{equation}{eq.}{}
\crefname{appendix}{Appendix}{}
\crefformat{section}{\S#2#1#3}  

\newcommand{\vtheta}{\boldsymbol \theta}
\newcommand{\vphi}{\boldsymbol \phi}
\newcommand{\ptheta}{p_{\vtheta}}
\newcommand{\qphi}{q_{\vphi}}
\newcommand{\xx}{\mathbf{x}}
\newcommand{\yy}{\mathbf{y}}
\newcommand{\EOS}{\texttt{EOS}\xspace}
\newcommand{\Sigmax}{\Sigma_{{\texttt{x}}}}
\newcommand{\Sigmay}{\Sigma_{{\texttt{y}}}}

\newcommand{\bitext}{{\cal B}}
\newcommand{\monotext}{{\cal M}}
\newcommand{\bitextback}{{\cal B}_\textit{back}}
\newcommand{\bitextdreamt}{{\cal B}_\textit{dreamt}}

\usepackage{xcolor}
\usepackage[disable]{todonotes}   
\newcommand{\note}[4][]{\todo[author=#2,color=#3,size=\scriptsize,fancyline,caption={},#1]{#4}} 
\newcommand{\ryan}[2][]{\note[#1]{ryan}{purple!40}{#2}}

\newcommand{\Ryan}[2][]{\ryan[inline,#1]{#2}\noindent}

\newcommand{\cutforspace}[1]{}

\title{Explaining and Generalizing Back-Translation through Wake-Sleep}
\author{
Ryan Cotterell \\
  Department of Computer Science\\
  Johns Hopkins University\\
  {\tt ryan.cotterell@jhu.edu} \\\And
  Julia Kreutzer \\
  Department of Computational Linguistics\\
  Heidelberg University \\
  {\tt kreutzer@cl.uni-heidelberg.de} \\}

\date{}

\begin{document}
\maketitle
\begin{abstract}
  Back-translation has become a commonly employed heuristic for
  semi-supervised neural machine translation. The
  technique is both straightforward to apply and has led to state-of-the-art
  results. In this work, we offer a principled interpretation of
  back-translation as approximate inference in a generative model of
  bitext and show how the standard implementation of back-translation
  corresponds to a single iteration of the wake-sleep algorithm
  in our proposed model. Moreover, this interpretation suggests a
  natural iterative generalization, which we demonstrate
  leads to further improvement of up to 1.6 BLEU.
\end{abstract}

\section{Introduction}
Recurrent neural networks have asserted hegemony in machine translation research.  Whereas
phrase-based machine
translation systems consisted of a hodgepodge of individual
components, expertly crocheted together to produce a final translation,
neural machine translation (NMT) is a fully end-to-end system
that treats learning a translator as parameter estimation in a single
discriminative probability model. To train
an NMT model in a semi-supervised fashion, an
interesting heuristic has emerged---back-translation, a simple
technique that hallucinates additional bitext from monolingual
data. In this work, we interpret and generalize back-translation using techniques from generative modeling and variational inference.

In broad strokes, back-translation works as follows. The NMT
practitioner trains \emph{two} systems: a forward translation system
that translates from the source to the target language and a second,
backwards translation system that translates from the target
back into the source language. The backwards translation system is
then used to translate additional monolingual text into the source
language, hallucinating, as it were, more bitext. Then, the additional data is used to estimate a higher-quality
forward translation model. Back-translation has recently risen to
fame in the context of NMT \cite{sennrich-haddow-birch:2016:WMT}
and has helped these systems achieve state-of-the-art results; the method's provenance, however, is older: \newcite{bertoldi-federico:2009:WMT}
and \newcite{li-EtAl:2011:EMNLP1} both applied
back-translation to non-neural MT. 

Our contribution is a novel interpretation and straight-forward extension of back-translation that rest on the
construction of a fully generative model of bitext. Working within this model,
we cast back-translation as a variational approximation, where the
backwards translator is an inference network that approximates a
posterior of a latent variable---the unobserved source sentence.
Specifically, we show that back-translation is a \emph{single
  iteration} of the wake-sleep variational scheme \cite{hinton1995wake}; this
interpretation suggests a simple extension to the model, where we
iteratively re-estimate both the forward and backward translator in a fashion
similar to expectation maximization.
We experiment on on two language pairs (English$\leftrightarrow$German and English$\leftrightarrow$Latvian) on two domains (WMT news translation and TED talks) and
find that our extension brings consistent gain over vanilla back-translation, up to 1.6 BLEU.

We may summarize
our paper concisely with the following koan: If back-translation is the answer, what was the question?

\section{A Generative Model of Bitext}
We construct a generative latent-variable model for the production of
bitext with the goal of showing that back-translation corresponds to a
form variational inference in the model. First, however, we establish
the requisite notation. Let $\Sigmax$ and $\Sigmay$ be finite
alphabets (of words)
for the source and target languages, respectively. Both are augmented with
a distinguished end-of-sentence symbol \EOS. Let $\xx \in \Sigmax^*$
and $\yy \in \Sigmay^*$ be strings, each of which ends with \EOS.
Formally, then, a monotext $\monotext$ is a collection
of sentences $\big\{\yy^{(i)}\big\}_{i=1}^N$  and a bitext is a
collection of \emph{aligned pairs} of sentences $\bitext = \big\{\langle
\xx^{(i)}, \yy^{(i)} \rangle \big\}_{i=1}^N$, where each $\yy^{(i)}$,
a sentence in the target language, is a translation of $\xx^{(i)}$, a
sentence in the source language.

We define then our generative model of bitext as
\begin{equation}\label{eq:joint}
p(\bitext) = \!\!\! \prod_{\langle \yy, \xx \rangle \in \bitext}\!\!\!\!
p(\yy, \xx)= \!\!\! \prod_{\langle \yy, \xx \rangle \in \bitext} \!\!\!\!
\ptheta(\yy \mid \xx)\, p(\xx)
\end{equation}
The distribution may be viewed as a directed graphical model (a
Bayesian network).
We will term the
model $\ptheta(\yy \mid \xx)$ the \textbf{translation model} and
$p(\xx)$ the \textbf{language model}.\cutforspace{\footnote{Warning: both of these
  terms are also used in noisy channel models for machine
  translation. We, however, define them slightly differently.}}  In
general, both $\ptheta(\yy \mid \xx)$ and $p(\xx)$ will be richly parameterized, such
as by a recurrent neural network---see \cref{sec:experiments}. As our
end task is machine translation, the distribution $\ptheta(\yy \mid
\xx)$ is the final product---we will discard $p(\xx)$. 

\paragraph{Supervised Machine Translation.}
Most machine translation models are estimated in the fully supervised
setting: one directly estimates the distribution
$\ptheta(\yy \mid \xx)$ through maximum likelihood estimation, i.e, maximizing
$\log p(\bitext) = \sum_{\langle \xx, \yy \rangle} \log \ptheta(\yy \mid
\xx)$. As $\ptheta(\yy \mid \xx)$ is often a continuous function of its parameters
$\vtheta$, gradient-based methods are typically employed.

\paragraph{Semi-Supervised Generalization.}
Lamentably, bitext in the wild is a relatively rare find, but monolingual text abounds.
A natural question is, then, how can we exploit this monolingual text in the estimation
of machine translation systems? Generative modeling provides the answer---we may optimize
the marginal likelihood, where we marginalize out the translation of the unannotated source
sentence; formally, this yields the following:
\begin{equation}
\prod_{\yy \in \monotext} p(\yy) = \prod_{\yy \in \monotext} \sum_{\xx \in \Sigmax^*} \ptheta(\yy \mid \xx)\, p(\xx) \label{eq:marginal}
\end{equation}
Unfortunately, there are an infinite
number of summands, which makes \cref{eq:marginal} intractable to compute.
Thus, we rely on an approximate strategy, one iteration of which of which will be shown to be equivalent to the back-translation technique. 

\section{Variational Back-Translation}\label{sec:inference}
\ryan{Semi-supervised learning is not discussed?}
Semi-supervised learning in the model requires efficient marginal inference.
To cope, we derive an approximation scheme, based on
the wake-sleep algorithm. Wake-sleep, originally presented in the context of the Helmholtz machine \cite{dayan1995helmholtz}, is an an iterative procedure that, prima facie, resembles the
expectation maximization (EM) algorithm of
\newcite{DemLaiRub77Maximum}. 
 Much like EM, wake-sleep has two steps that are to be alternated: (i)
the sleep phase and (ii) the wake phase.

\subsection{Overview}
Before discussing the algorithmic details, we give the intuition behind
the connection we draw. Wake-sleep will iterate between learning a better forward-translator
$\ptheta(\yy \mid \xx)$ and a better back-translator $\qphi(\xx \mid \yy)$.
Typically in back-translation, however, the back-translator is trained and
then additional bitext is hallucinated \emph{once} in order to train a better forward-translator
$\ptheta(\yy \mid \xx)$. However, under the view that $\qphi(\xx \mid \yy)$ should be an
approximation to the posterior $p(\xx \mid \yy)$ in the joint model \cref{eq:joint},
the wake-sleep algorithm suggests
and iterative procedure that gradually refines $\qphi$, taking information from $p$ into account.
Thus, under wake-sleep, we constantly retrain the forward-translator $\ptheta(\yy \mid \xx)$
using the updated back-translator and vice versa. 

\begin{algorithm}[t]
  \begin{algorithmic}[1]
  	\Require{initial forward \& backward NMT parameters $\vtheta$, $\vphi$; monotext $\monotext$; language model $p(\cdot)$ }
  	\Ensure{final model parameters $\vtheta$, $\vphi$ }
    \For{$i=1$ \textbf{to} $I$}
    \State $\bitextback \gets \emptyset$
    \For{$\yy \in \monotext$}
    	\State $\tilde{\xx} \sim \qphi(\cdot \mid \yy)$
    	\State $\bitextback \gets \bitextback \cup \{\langle \yy, \tilde{\xx} \rangle \}$
    \EndFor
    \State {estimate} $\vtheta$ {by} {maximizing} $\log \ptheta$ of $\bitext \,\cup\,\bitextback$
    \State $\bitextdreamt \gets \emptyset$
    \For{ $\tilde{\xx} \sim p(\cdot)$ }
    \State $\tilde{\yy} \sim \ptheta(\cdot \mid \tilde{\xx})$
    \State $\bitextdreamt \gets \bitextdreamt \cup \{ \langle \tilde{\yy}, \tilde{\xx}\rangle \}$
    \EndFor
     \State { estimate} $\vphi$ { by maximizing}  $\log \qphi$ {of} $\bitext 	\,\cup\,\bitextdreamt$    \EndFor
\end{algorithmic}
  \caption{{Wake-sleep for Semi-Supervised Neural Machine Translation}}
  \label{alg:wake-sleep}
\end{algorithm}

\subsection{The Sleep Phase}
Had we access to the true posterior of our joint model $p(\xx \mid
\yy)$, we could apply EM in a straightforward manner, as one does in
models that admit tractable computation of the quantity, e.g., the
hidden Markov model \cite{rabiner1989tutorial}. However, in general we will choose a
rich neural parameterization that will prohibit its direct computation; thus, we seek a distribution $\qphi(\xx \mid
\yy)$ that well-approximates $p(\xx \mid \yy)$.
What is $\qphi(\xx \mid \yy)$? In the machine learning literature, this
distribution is termed an \textbf{inference network}---a parameterized
distribution that approximate the posterior over $\xx$ for \emph{any}
observed sentence $\yy$. Inference networks have been applied to a wide variety of problems, e.g., topic modeling \cite{miao2016neural}
and inflection generation \cite{zhou-neubig:2017:Long}.\footnote{While
  both of these approaches employ inference networks to reconstruct the latent variables,
  neither makes use of the wake-sleep procedure.}

A common principled manner to approximate a probability distribution is to
minimize the Kullback-Leibler (KL) divergence. The sleep step dictates
that we choose $\qphi$ so as to minimize the quantity
\begin{equation}
\sum_{\yy \in \monotext}  D_\textit{KL}\Big(p(\cdot \mid \yy) \mid\mid \qphi(\cdot \mid \yy) \Big) \label{eq:kl}
\end{equation}
These inclusive KL divergences are still intractable---we would have to normalize
the distribution $\ptheta(\xx \mid \yy)$, which is hard since it involves
as sum over $\Sigmax^*$.
By design, however, our model $p(\yy, \xx)$ is a directed generative
model so we can efficiently generate samples through forward sampling;
first, we sample a sentence $\tilde{\xx}^{(i)} \sim p(\cdot)$ and then
we sample its translation $\tilde{\yy}^{(i)} \sim \ptheta(\cdot \mid
\tilde{\xx}^{(i)})$. 
We term the new bitext of $M$ samples the
``dreamt'' bitext $\bitextdreamt = \{ \langle \tilde{\yy}^{(i)},
\tilde{\xx}^{(i)} \rangle \}_{i=1}^M$. Using these samples from the joint $p(\yy, \xx)$,
we may approximate the true posterior $p(\yy \mid \xx)$ by maximizing
the following Monte Carlo approximation to \cref{eq:kl}: $\sum_{ \langle \tilde{\yy},
  \tilde{\xx} \rangle \in \bitextdreamt} \log \qphi(\tilde{\xx} \mid
\tilde{\yy})$. To find good parameters $\vphi$, we will optimize
the log-likelihood of $\bitext\,\cup\,\bitextdreamt$ through a gradient-based method, such
as backpropagation \cite{rumelhart1985learning}.

\subsection{The Wake Phase}
Equipped with our approximate posterior $\qphi(\xx \mid \yy)$, the
wake phase proceeds as follows. For every observed sentence $\yy \in
\monotext$, we sample a back-translation $\tilde{\xx} \sim \qphi(\cdot
\mid \yy)$, creating a bitextual extension $\bitextback = \big\{ \yy,
\tilde{\xx} \big\}_{\yy \in \monotext}$ of the monotext
$\monotext$. Now, we may train full joint model $p(\yy, \xx)$ using
both the original bitext and the sampled bitext, i.e., we train on
$\bitext \,\cup\, \bitextback$, in the fully supervised setting. More
concretely, in the wake phase we train the model parameters $\vtheta$
with backpropagation. Both steps are alternated, as in EM, until
convergence.  The pseudocode for the full procedure is given in
\cref{alg:wake-sleep}.  The procedure also resembles variational EM
\cite{beal2003variational}. The difference is that wake step
minimizes an inclusive KL, rather than the exclusive one
found in variational EM.

\begin{figure}[t]
\centering
\includegraphics[width=\columnwidth]{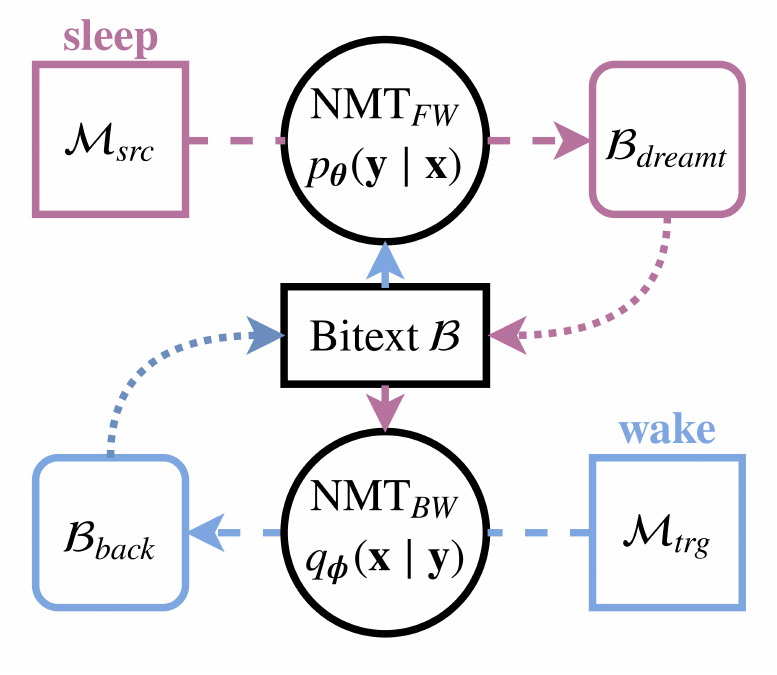}
\caption{Diagram for semi-supervised back-translation with $\text{NMT}_{\textit{FW}}$ translating $\textit{src}\rightarrow\textit{trg}$, $\text{NMT}_{\textit{BW}}$ translating $\textit{trg}\rightarrow\textit{src}$, two monotexts $\cal{M}_{\textit{src}}$, $\cal{M}_{\textit{trg}}$ and a bitext $\bitext$. Solid arrows indicate training, dashed translation and dotted a union.}
\label{fig:diagram}
\end{figure}

\paragraph{Implicitly Defining the Language Model.}
We are uninterested in the language model
$p(\xx)$---we only require it in order to generate samples for
the sleep phase. Thus, rather than taking the
time to estimate a language model $p(\xx)$ and to sample from it, which would almost
certainly be of lower quality than additional monolingual text, we
simply randomly sample existing sentences from a large monolingual
corpus in the source language. Note that this corresponds to defining
$p(\xx)$ to be a categorical distribution over entire sentences
attested in the monotext considered.

\subsection{Interpretations and Insight}\label{sec:interpret}
\paragraph{Interpretation as Back-translation.}
One iteration ($I=1$) of the algorithm described in
\cref{alg:wake-sleep}: we first train a back-translator and then
annotate monolingual data to improve $\ptheta$. Note that the dream
phase is irrelevant here. One difference that is worth noting is that
in \cref{alg:wake-sleep}, additional forms are \emph{sampled}, whereas
many attempt a one-best decode to get back-translations.  We may simply view the (approximation) maximization as a Viterbi
approximation to the expectation, as justified by
\newcite{nh-nveajis-98}. The back-translation algorithm of
\newcite{sennrich-haddow-birch:2016:P16-11} is best termed
one iteration of Viterbi wake-sleep, illustrated in \cref{fig:diagram}.
\paragraph{Interpretation as an Autoencoder.}
As \newcite{kingma2013auto} saw, we can
alternatively view the relation between the original model
$p_{\vtheta}$ and the inference network $\qphi$ as an
autoencoder. Specifically, we may think this procedure as optimizing
the autoencoding objective: $\sum_{\xx \in \Sigmax^*}
  \ptheta(\yy \mid \xx) \qphi(\xx \mid \yy)$.
where we have temporarily neglected the prior $p(\xx)$. Indeed,
this is an interesting autoencoder as our latent variable
is structured, $\Sigmax^*$, rather than $\mathbb{R}^n$, as in \newcite{kingma2013auto}. Such structure suggests a relation
to the conditional random field autoencoder of \newcite{ammar2014conditional}.

\begin{table*}
  \centering
  \begin{tabular}{l|ll|ll|ll} 
  \toprule
     & \multicolumn{2}{c|}{\textbf{TED}} & \multicolumn{4}{c}{\textbf{WMT 2017}} \\ 
     & \textbf{en-de} & \textbf{de-en} & \textbf{en-de} & \textbf{de-en} & \textbf{en-lv} & \textbf{lv-en} \\ \midrule
Iteration 0  &   24.38   &  27.29      &   20.73 &  25.41  & 11.41 &  12.53 \\
Iteration 1  &   25.58$^{\star\dagger}$  &   29.80$^{\star\dagger}$     &    21.63$^{\star\dagger}$ &  26.63$^{\star\dagger}$ &  12.76$^{\star\dagger}$ &  12.42 \\
Iteration 2   &  26.73$^{\star\dagger}$  &   30.02$^{\star\dagger}$     &    \textbf{22.33}$^{\star\dagger}$ &  \textbf{26.80}$^{\dagger}$ &  \textbf{12.91}$^{\dagger}$ &  13.43$^{\star\dagger}$ \\
Iteration 3   &  \textbf{27.20}$^{\star\dagger}$  &   \textbf{30.21}$^{\star\dagger}$    &     21.72$^{\star\dagger}$ &  26.26$^{\star\dagger}$ &  12.77$^{\star\dagger}$ &  \textbf{13.53}$^{\star}$ \\
\midrule
$\Delta(\text{best, Iteration 1})$ & +1.62  &  +0.41   &  +0.70   & +0.17   & +0.15 & +1.10  \\
$\Delta(\text{best, Iteration 0})$ & +2.82   &  +2.92    &     +1.59 &  +1.39  & +1.5 &   +1.0   \\
    \bottomrule
  \end{tabular} %
  \caption{Results on the TED and WMT 2017 test data as reported by \textsc{SacreBLEU} (TED: \texttt{BLEU+case.lc+numrefs.1+smooth.exp+tok.13a.version.1.2.3}, WMT: \texttt{BLEU+case.mixed+numrefs.1+smooth.exp+tok.13a.version.1.2.3}). Iteration 0 is the MLE-trained model without back-translation, Iterations 1-3 describe the models resulting from subsequent iterations of Algorithm \ref{alg:wake-sleep}.
  Significant differences (at $p < 0.05$) to the respective previous iteration are marked with `$\star$', significant differences to Iter 0 with `$\dagger$'. }
  \label{tab:results}
\end{table*}

\section{Related Work}\label{sec:related-work}
Back-translation as a technique for semi-supervised
machine translations dates to the phrase-based era; see \newcite{bertoldi-federico:2009:WMT}.  Interestingly, many techniques explored
in the context of phrase-based translation have yet to be
neuralization---consider \cite{li-EtAl:2011:EMNLP1}, who offered a minimum risk back-translation strategy.

The contemporary use of back-translation in state-of-the-art (e.g. \newcite{HumanParity18}) and unsupervised neural machine translation \cite{artetxe2018unsupervised} dates back to large empirical gains found by \newcite{sennrich-haddow-birch:2016:WMT} in
the context of neural MT and automatic post-editing \cite{junczysdowmunt-grundkiewicz:2016:WMT}.
Our work is distinguished from these previous papers in that we are
interested in a principled interpretation of back-translation as a
strategy, independent of the particular parameterization in
place; our analysis will hold parameterization of a
probabilistic MT model, e.g., \newcite{liang-EtAl:2006:COLACL} and \newcite{blunsom-osborne:2008:EMNLP}.
Moreover, our analysis suggests an iterative extension that we will show leads to better empirical performance in \cref{sec:experiments}.

Finally, our work is related to the dual learning method of
\newcite{he2016dual} who, like us, suggested an iterative approach to
back-translation. While spiritually related, the motivation for our
respective algorithms are quite different; they motivate their
procedure game-theoretically. Furthermore, translation models in both directions are updated with online reinforcement learning, after one batch of translations each. \ryan{Check that it is actually
  game-theoretic?}
\Ryan{
  \newcite{zhang2016variational} and \newcite{su2018variational} have proposed Bayesian
  methods for smoothing the parameter estimates of a NMT system and variational inference allows for approximate Bayesian computation. The motivation for this
  work is orthogonal to ours. }

\section{Experiments}\label{sec:experiments}
A core contribution of this paper is theoretical---we sought a
principled interpretation back-translation, which is commonly seen as
a heuristic. However, our analysis motivated an iterative extension to
the algorithm. Naturally, we will want to show that extension leads to better results.
Our experimental paradigm is, then, a controlled
comparison between the original back-translation method and the
wake-sleep extension. Note that our algorithm recovers the original back-translation method in the special case that we only run 1 iteration.

\begin{table}
\center
\begin{tabular}{lllll}
\toprule
\textbf{Domain} & \textbf{Language(s)} & \textbf{Train} & \textbf{Dev} & \textbf{Test}\\
\midrule
WMT & de$\leftrightarrow$en & 5.9M & 2999 & 3004 \\
WMT & lv$\leftrightarrow$en & 4.5M & 2003 & 2001\\
TED & de$\leftrightarrow$en & 153k & 6969 & 6750 \\
WMT & de & 500k & - & - \\
WMT & en & 500k & - & - \\
WMT & lv & 500k & - & - \\
\bottomrule
\end{tabular}
\caption{Number of sentences in mono- and bitexts used in the experiments. The WMT monotexts are selected randomly from the WMT news crawls.}
\label{tab:data}
\end{table}

\paragraph{Data.}
We consider translations from English to German, English to Latvian and vice versa and use the news translation WMT 2017 and TED data from IWSLT 2014 for our experiments. 
Pre-processed WMT17 data was provided by the official shared task.\footnote{\url{http://www.statmt.org/wmt17/translation-task.html}} Pre-processed data splits for TED were the same as in \cite{BahdanauETAL:17}.\footnote{Obtained from \url{https://github.com/rizar/actor-critic-public/tree/master/exp/ted}.}
Table \ref{tab:data} lists number of sentences for the data used in the experiments. \\
We investigate two scenarios: 1) standard back-translation with additional monolingual data from the same domain (WMT) and 2) back-translation for semi-supervised domain adaptation (TED). 
In both cases we start with standard supervised training on the WMT bitext. For WMT experiments, 500k sentences from monolingual news crawls provided in the shared task are randomly selected and used for back-translation iterations. For TED, each side of the training data (153k sentences) serves as monotext for semi-supervised domain adaptation via back-translation. 

\paragraph{Machine Translation Model.} We choose a classic recurrent encoder-decoder architecture with attention \cite{cho-EtAl:2014:SSST-8,DBLP:conf/nips/SutskeverVL14,DBLP:journals/corr/BahdanauCB14}. The NMT has a bidirectional encoder and a single-layer decoder with
1024 Gated Recurrent Unit \cite{cho-EtAl:2014:SSST-8} each, and
subword embeddings of size 500 for a shared vocabulary of subwords
obtained from 30k byte-pair merges
\cite{sennrich-haddow-birch:2016:P16-12}.
Maximum input and output sequence length are set to 60. 
The model parameters are optimized with Adam ($\alpha=10^{-4}$, $\beta_1=0.9$, $\beta_2=0.999$, $\epsilon=10^{-8}$) 
\cite{KingmaBa:14} on mini-batches of size 60. To prevent the models
from overfitting, dropout with probability 0.2
\cite{SrivastavaETAL:14} and l2-regularization with weight $10^{-8}$
are applied during training. The gradient is clipped when its norm
exceeds 1.0 \cite{PascanuETAL:13}. All models are trained on a maximum of 10 epochs on their respective training data.

\paragraph{Wake-Sleep.}
First, we train models for all directions with maximum likelihood estimation on the original WMT bitext. Then, each model translates the monolingual data to serve as back-translator for the opposite direction. Pairing the ``dreamt'' sources with the original targets, the models are further trained on the new bitext (see \cref{fig:diagram}). Early stopping points are
determined on the development set for each iteration. For back-translation we use greedy decoding (Viterbi approximation, see \cref{sec:interpret}),
for testing beam search of width 10.

\paragraph{Results and Discussion.}
The results may be found in \cref{tab:results}.  The models are
evaluated with respect to BLEU \cite{PapineniETAL:02} using the \textsc{SacreBLEU} tool (v.1.2.3) \cite{post2018call} on detokenized (WMT: recased) system outputs.\footnote{\url{https://github.com/awslabs/sockeye/tree/master/contrib/sacrebleu}}
When comparing the results between iterations, we observe the largest relative improvements in iteration 1, the original back-translation, with the exception of lv-en, where it is in iteration 2. Our iterative extension further improves over these results in all experiments. The gains are highest in the TED domain, since the back-translations enable adaptation to the new domain. For WMT the gains are smaller (and come earlier) since the monotext, bitext and test data originate from the same (or at least similar) domain. Despite the wide range of absolute BLEU scores, WMT en-lv being the weakest, TED de-en being the strongest model, back-translation iterations can in all cases achieve an overall improvements over at least 1 BLEU using relatively small amounts of monotexts.
Potentially larger gains could be achieved by leveraging more monolingual data, and by employing more sophisticated data selection strategies to filter out potential noise in the monotexts. 

\section{Conclusion}
We have provided a principled interpretation
and generalization
of back-translation as variational inference in a generative
model of bitext.
\Ryan{Specifically, we argued that back-translation
corresponds to an instantiation of the wake-sleep algorithm.
As wake-sleep is an iterative algorithm, our analysis has led to a simple
iterative generalization of back-translation.} Experimentally,
we have shown that this leads to improvements of up to 1.6 BLEU over a back-translation baseline. We believe that a cleaner
understanding of nature of back-translation will yield future
innovations and extensions and hope our attempt is a step in that direction. 

\bibliography{naaclhlt2018}
\bibliographystyle{acl_natbib_nourl}

\end{document}